\documentclass{ifacconf}

\usepackage{graphicx}      
\usepackage{natbib}        
\usepackage{amsmath,amsfonts,amssymb}
\usepackage{tikz}
\usepackage{pgfplots}
\usepackage{url}

\definecolor{color1}{rgb}{0, 0.4470, 0.7410}
\definecolor{color2}{rgb}{0.8500, 0.3250, 0.0980}
\definecolor{color3}{rgb}{0.4660, 0.6740, 0.1880}
\tikzset{plotline/.style={line width=0.9}}
\tikzset{plotline1/.style={plotline, color=color1, mark=square}}
\tikzset{plotline2/.style={plotline, color=color2, mark=*}}
\tikzset{plotline3/.style={plotline, color=color3, mark=triangle}}

\begin{document}
\begin{frontmatter}

\title{Multiagent Rollout with Reshuffling for Warehouse Robots Path Planning\thanksref{footnoteinfo}} 

\thanks[footnoteinfo]{This work was supported by the Swedish Foundation for Strategic Research, and the Swedish Research
Council.}

\author[First]{William Emanuelsson} 
\author[First]{Alejandro Penacho Riveiros} 
\author[First]{Yuchao Li}
\author[First]{Karl H. Johansson}
\author[First]{Jonas M\aa rtensson}

\address[First]{Division of Decision and Control Systems, KTH Royal Institute of Technology, Sweden, (e-mail: wem,alejpr,yuchao,kallej,jonas1@kth.se).}

\begin{abstract}                
Efficiently solving path planning problems for a large number of robots is critical to the successful operation of modern warehouses. The existing approaches adopt classical shortest path algorithms to plan in environments whose cells are associated with both space and time in order to avoid collision between robots. In this work, we achieve the same goal by means of simulation in a smaller static environment. Built upon the new framework introduced in \citep{bertsekas2021multiagent}, we propose multiagent rollout with reshuffling algorithm, and apply it to address the warehouse robots path planning problem. The proposed scheme has a solid theoretical guarantee and exhibits consistent performance in our numerical studies. Moreover, it inherits from the generic rollout methods the ability to adapt to a changing environment by online replanning, which we demonstrate through examples where some robots malfunction.

\end{abstract}


\end{frontmatter}

\section{Introduction}\label{sec:intro}
In recent years, the trade of goods over the internet, commonly referred to as `e-commerce,' has been consistently increasing its influence due to its convenience to consumers and the large variety of options available. Companies involved in this business model have experienced massive growth, but must also manage increasingly complex logistical infrastructures. Among the most critical parts of them are fulfillment centers. These are massive warehouses tasked with the storage and handling of a large number of goods at the same time. The efforts to automatize these centers must deal with the challenge of coordinating hundreds of robots to transport items inside the warehouse efficiently. The coordination problem is typically decomposed into several sub-problems [see, e.g., \citep{wurman2008coordinating}], and in this work, we focus on path planning problem, which can be modeled adequately as computing shortest paths for a large number of robots in a grid environment, as in, for example, \citep{li_lifelong_2021}. 

Research on shortest path problems has a long and rich history. Some of the most well-established algorithms include Bellman-Ford \citep{bellman_routing_1958}, Dijkstra \citep{dijkstra_note_1959}, A* \citep{hart_formal_1968} and Auction \citep{bertsekas1991auction}. These algorithms exploit different properties of the problem to minimize the computational cost, and collectively address the problem sufficiently well. However, they only tackle problems involving a single agent, thus making them not directly applicable to warehouse path planning. More recent efforts have built upon previous methods to tackle problems with several agents. Algorithms of this kind include cooperative A* \citep{silver_cooperative_2005}, M* \citep{wagner_m_2011} and conflict based search \citep{sharon_conflict-based_2015}. They offer different possibilities in the trade-off between computational cost and optimality for multiagent path finding. The common feature of those approaches is to augment time to space and create an enlarged grid environment, with each cell associated with both physical space and time. By searching through this enlarged environment, agents can avoid collisions between each other.

Still, when considering the path planning problem in a warehouse, the aforementioned algorithms are not readily applicable. This is because they only deal with one-shot problems, where each agent only has to move from its initial position to a single target position. In contrast, in a real warehouse, a robot is expected to be assigned endless tasks, picking and delivering items one after the other. In other words, it is more suitable to regard the problem as one involving \emph{infinite horizon}, i.e., planning infinitely long paths. Some recent works have designed algorithms that are built upon the previous results and are tailored for this feature, and \citep{ma_lifelong_2017} and \citep{li_lifelong_2021} are notable successes among them.    

Parallel to the algorithmic progress discussed above, some high-profile successes in the field of reinforcement learning (RL), such as \citep{silver2018general}, have taken the world by storm. Among the tools offered by RL, a broad class of algorithms is couched on the theory of dynamic programming, which in turn laid the foundation for the Bellman-Ford algorithm. Therefore, it is no surprise that tools developed in RL can be brought to bear to address the problem considered here. The key distinguishing factor of RL methods from those discussed above is its use of a simulator, where simulations of future results are applied to facilitate decision making at the current stage. In particular, through the use of simulation in a smaller grid environment than those applied in the above schemes, it is possible to detect potential collisions between agents, thus leading to proper decisions at the present stage. For a textbook account of the subject, see \citep{sutton2018reinforcement,bertsekas2019reinforcement}. 

Among a great variety of methods in RL, \emph{multiagent rollout} is well-suited for the warehouse path planning problem. Like the rollout method first developed in \citep{tesauro1996line} for the game of backgammon, it relies on real-time simulation results of a known heuristic for decision making. An appealing consequence of this nature is its strong ability to adapt to a changing environment. The additional feature that is particularly tailored for multiagent problems is that it focuses on the decision selection process one agent at a time, thus avoiding the computational burden rendered by large decision spaces. The method was outlined in \citep[Section~6.1.4]{bertsekas1996neuro}, and has been substantially extended in \citep{bertsekas2021multiagent}. Further details of the method will be given in the next section.

In this work, we build upon the multiagent rollout algorithm, and propose a variant of it that involves random reshuffling of the orders of agents. The proposed method is applied to warehouse robot path planning problems, where up to $200$ robots are involved. In our numerical study, the proposed scheme demonstrates a better rate of success compared with both standard rollout, as well as a scheme adapted from cooperative A*. With suitable pre-computation of shortest paths for a static environment, the average computation time can be as little as $50$ ms. 

The remainder of the paper is organized as follows. Section~\ref{sec:multi_rollout} introduces the mathematical problem, and describes the proposed scheme for this problem. In Section~\ref{sec:model}, we provide further details on the modeling of warehouse path finding. In Section~\ref{sec:numerical} we demonstrate the results of our numerical studies, where our approach is compared with standard multiagent rollout as well as a variant of cooperative A*.

\section{Multiagent Rollout with reshuffling}\label{sec:multi_rollout}
In this section, we describe the multiagent rollout with reshuffling algorithm, which is later applied to plan paths for warehouse robots. We will first introduce the multiagent optimal control problems with which we model the path planning problem of our interest, and then provide details of the rollout method. Our presentation is somewhat abstract, leaving further details specific to warehouse path planning in the next section.

\subsection{Multiagent Deterministic Optimal Control}
The problem considered here involves stationary dynamics
\begin{equation}
\label{eq:dynamics}
    x_{k+1}=f(x_k,u_k),\quad k=0,\,1,\,\dots,
\end{equation}
where $x_k$ and $u_k$ are state and control at stage $k$, which belong to state and control spaces $X$ and $U$, respectively, and $f$ maps $X\times U$ to $X$. Both the state and control spaces contain finitely many elements. The control $u_k$ must be chosen from a nonempty constraint set $U(x_k)\subset U$ that may depend on $x_k$. The cost of applying $u_k$ at state $x_k$ is denoted by $g(x_k,u_k)$, and is assumed to be real-valued:
\begin{equation}
\label{eq:cost}
    g(x_k,u_k)\in \Re,
\end{equation}
where $\Re$ denotes the real line. We consider feedback policy $\mu$ that maps $X$ to $U$ and satisfying $\mu(x)\in U(x)$ for all $x$.

The \emph{cost function} of a policy $\mu$, denoted by $J_\mu$, maps $X$ to $\Re$, and is defined at any initial state $x_0 \in X$, as
\begin{equation}
\label{eq:pi_cost_function}
    J_\mu(x_0)= \sum_{k=0}^\infty \alpha^kg(x_k,\mu(x_k)),
\end{equation}
where $\alpha\in (0,1)$ is called \emph{discount factor}, which reflects the emphasis on the current cost over future ones, and $x_{k+1}=f(x_k,\mu(x_k))$, $k=0,\,1,\,\dots$. The limit in \eqref{eq:pi_cost_function} is well-posed in view of the finiteness of $X$ and $U$. The optimal cost function $J^*$ is defined pointwise as
\begin{equation}
    J^*(x_0)=\inf_{\substack{u_k\in U(x_k),\ k=0,1,\ldots\\ x_{k+1}=f(x_k,u_k),\ k=0,1,\ldots}}\sum_{k=0}^\infty  \alpha^kg(x_k,u_k).
\end{equation}
A stationary policy $\mu^*$ is called optimal if
\begin{equation*}
    J_{\mu^*}(x)=J^*(x),\quad \forall x\in X.
\end{equation*}

The multiagent nature of the problem manifests itself through the Cartesian product structure of $U(x_k)$, i.e.,
\begin{equation}
    \label{eq:cartesian}
    U(x_k)=U^1(x_k)\times \cdots \times U^m(x_k),
\end{equation}
where $m$ is an integer representing the number of agents involved in the problem. As a result, $u_k$ can be written as
$$u_k=(u_k^1,\dots,u_k^m),$$
with $u_k^i\in U^i(x_k)$, $i=1,\dots,m$. Accordingly, the system dynamics \eqref{eq:dynamics} and stage cost \eqref{eq:cost} can be written as
$$x_{k+1}=f(x_k,u^1_k,\dots,u^m_k),\quad g(x_k,u^1_k,\dots,u^m_k),$$
respectively, with $u_k=(u^1_k,\dots,u^m_k)$ when stressing the multiagent structure of the problem. In addition, policy itself $\mu$ can be decomposed similarly as
$$\mu(x)=\big(\mu^1(x),\dots,\mu^m(x)\big),$$
with $\mu^i(x)\in U^i(x)$, $i=1,\dots,m$.

For the problem considered here, it can be shown that there exists a stationary optimal policy $\mu^*$; see, i.e., \citep[Props.~4.3.2 and 4.3.4]{bertsekas2019reinforcement}. Two exact DP algorithms for addressing the problem are value iteration (VI) and policy iteration (PI). However, these algorithms may be intractable, due to the \emph{curse of dimensionality}, which refers to the explosion of the computation as the cardinality of $X$ increases. Thus, various approximation schemes have been proposed, where one aims to obtain some suboptimal policy $\Tilde{\mu}$ such that $J_{\Tilde{\mu}}\approx J^*$. These schemes often involve minimizing some objective over the set $U(x)$. Yet, due to the structure \eqref{eq:cartesian} of $U(x)$, even those suboptimal schemes can be rather challenging as searching through the set $U(x)$ can be too costly as well. To see this, suppose each set $U^i(x)$ contains at most $C$ elements, then the total number of elements in $U(x)$ can be as large as $C^m$, which is prohibitively expensive to search through even for modest $m$. The multiagent rollout algorithm proposed in \citep{bertsekas2021multiagent} addresses exactly this problem, as we will discuss next.   

\subsection{Multiagent Rollout}
The standard rollout scheme relies on the use of a policy, which is called \emph{base policy}. It improves upon the base policy by searching through the set $U(x)$ in real-time, and produces a new policy which is called the \emph{rollout policy}. It can be shown that the rollout policy is guaranteed to outperform the base policy in the sense that its cost function values are no more than that of the base policy at every state; see \citep[Prop.~4.6.1]{bertsekas2019reinforcement}.

In particular, given a base policy $\mu$ and the current state $x$, the rollout policy is defined by performing minimization
\begin{equation}
    \label{eq:standard_rollout}
    \min_{u\in U(x)}\big\{g(x,u)+\alpha J_\mu\big(f(x,u)\big)\big\}
\end{equation}
in real-time and applies the control that attains the minimum. Since the cost function of the base policy $\mu$ is typically not in closed form, obtaining their values involves certain forms of real-time computation as well, such as simulation. Therefore, the above minimization requires computing as many values $J_\mu\big(f(x,u)\big)$ as the number of elements in $U(x)$, and comparing just as many numbers in the form of \eqref{eq:standard_rollout}. In the case of a multiagent system, the computation amount can be daunting in face of hard time constraints, which makes the scheme impractical.

Multiagent rollout circumvents the challenge by improving the policy \emph{one-agent-at-a-time}. Specifically, given the base policy $\mu=(\mu^1,\dots,\mu^m)$, it performs a sequence of minimization
\begin{equation}
    \label{eq:multi_rollout}
    \begin{aligned}
        \Tilde{\mu}^1(x)&\in\arg\min_{u^1\in U^1(x)}\Big\{g\big(x,u^1,\mu^2(x),\dots,\mu^m(x)\big)\\
    &+\alpha J_\mu\Big(f\big(x,u^1,\mu^2(x),\dots,\mu^m(x)\big)\Big)\Big\},\\
    \Tilde{\mu}^2(x)&\in\arg\min_{u^2\in U^2(x)}\Big\{g\big(x,\Tilde{\mu}^1(x),u^2,\mu^3(x),\dots,\mu^m(x)\big)\\
    &+\alpha J_\mu\Big(f\big(x,\Tilde{\mu}^1(x),u^2,\mu^3(x),\dots,\mu^m(x)\big)\Big\},\\
    \dots& \qquad \dots\qquad \dots\\
    \Tilde{\mu}^m(x)&\in\arg\min_{u^m\in U^m(x)}\Big\{g\big(x,\Tilde{\mu}^1(x),\dots,\Tilde{\mu}^{m-1}(x),u^m\big)\\
    &+\alpha J_\mu\Big(f\big(x,\Tilde{\mu}^1(x),\dots,\Tilde{\mu}^{m-1}(x),u^m\big)\Big)\Big\},
    \end{aligned}
\end{equation}
and applies the control $\Tilde{\mu}(x)$ that is defined as
$$\Tilde{\mu}(x)=\big(\Tilde{\mu}^1(x),\dots,\Tilde{\mu}^m(x)\big).$$

Let us assume that each set $U^i(x)$ contains at most $C$ elements as before. Then the above minimization involves evaluating and comparing at most $Cm$ values, which is a substantial improvement compared with $C^m$ values needed in \eqref{eq:standard_rollout}. In addition, if we denote the optimal value at the end of the minimization as $\Tilde{J}(x)$, i.e., 
\begin{align*}
    \Tilde{J}(x)=&\min_{u^m\in U^m(x)}\Big\{g\big(x,\Tilde{\mu}^1(x),\dots,\Tilde{\mu}^{m-1}(x),u^m\big)\\
    &+\alpha J_\mu\Big(f\big(x,\Tilde{\mu}^1(x),\dots,\Tilde{\mu}^{m-1}(x),u^m\big)\Big)\Big\},
\end{align*}
or equivalently 
\begin{equation}
\label{eq:j_tilde}
    \Tilde{J}(x)=g\big(x,\Tilde{\mu}(x)\big)+\alpha J_\mu\Big(f\big(x,\Tilde{\mu}(x)\big)\Big),
\end{equation}
then we can show that 
\begin{equation}
    \label{eq:cost_bound}
    J_{\Tilde{\mu}}(x)\leq \Tilde{J}(x),\quad \forall x\in X.
\end{equation}
A formal statement as well as its proof is given in Appendix~\ref{app:bound}. Besides, it is shown in \citep[Prop.~2]{bertsekas2021multiagent} that the multiagent rollout policy $\Tilde{\mu}$ outperforms the base policy in the same sense as in the standard rollout case, i.e.,
\begin{equation}
    \label{eq:rollout_improve}
    J_{\Tilde{\mu}}(x)\leq J_\mu(x),\quad \forall x\in X.
\end{equation}
In fact, rollout and multiagent rollout can be regarded as one step of Newton's method for solving the original optimal control problem and its equivalent reformulation, respectively, as shown in \citep[Section~3.4 and 3.7]{bertsekas2021multiagent}. Therefore, the produced policies are likely to be much better than the base policy. However, it is not known in general whether the standard rollout policy is better than the multiagent one, despite the much more effort being required. Empirical studies given in \citep{bhattacharya2021multiagent} addressing a partial information problem have found that these two have quite comparative performance.

Owing to its flexibility, multiagent rollout admits various forms of modifications and extensions. Extensive discussions can be found in \citep[Sections~3.2, 5.3 and 5.6]{bertsekas2021rollout}. One extension that is essential for our application involves multiple base policies. In particular, suppose that we have access to policies $\mu_1$ and $\mu_2$. This form of rollout defines a function $\Bar{J}:X\mapsto \Re$ as $\Bar{J}(x)=\min\{J_{\mu_1}(x),J_{\mu_2}(x)\}$, and replaces $J_\mu$ with $\Bar{J}$ in the sequence of minimization \eqref{eq:multi_rollout}. It can be shown that the rollout policy $\Tilde{\mu}$ is better than both $\mu_1$ and $\mu_2$, i.e.,
$$J_{\Tilde{\mu}}(x)\leq J_{\mu_1}(x),\quad J_{\Tilde{\mu}}(x)\leq J_{\mu_2}(x),\quad \forall x\in X.$$
In addition, the bound \eqref{eq:cost_bound} also holds, with $J_\mu$ replaced by $\Bar{J}$ in \eqref{eq:j_tilde} which defines the corresponding bound $\Bar{J}$.

\subsection{Multiagent Rollout with Reshuffling}
The multiagent rollout framework offers many options for further improvement. Among them, we focus on the order of agents according to which the minimization similar to \eqref{eq:multi_rollout} is carried out. Ideas on optimizing orders for similar optimization problems have appeared in the context of, for example, path planning of robots, under the name of `prioritized planning;' see \citep[Section~2.3.1]{latombe2012robot} and \citep{vcap2015prioritized}. In this work, however, we will rely on orders that are generated randomly. Before getting into the details of our scheme, we first note that the performance bound \eqref{eq:cost_bound} remains valid even if the agent-by-agent optimization is performed in a different order other than the default one. 

The basis of our algorithm is an assumption that we know a `rule' that provides a verdict on whether a control $u\in U(x)$ at a given state $x$ is `good.' We introduce permutation functions, denoted by $\sigma$ and $\tau$, which are one-to-one mappings with both domain and image being $\{1,\dots,m\}$. Thus, their inverse images exist and are denoted as $\sigma^{-1}$ and $\tau^{-1}$ respectively. 

Given the current state $x$ and permutation $\sigma$, we define $\tau_0=\sigma$. In addition, we define $\ell_0^i$ as
\begin{equation}
\label{eq:ell_tau}
    \ell_0^i=\tau_0^{-1}(i),\quad i=1,\dots,m.
\end{equation}
Then with a slight abuse of notation, we rearrange the elements of control according to $\tau_0$, i.e.,
$$u=\Big(u^{\ell_0^1},u^{\ell_0^2},\dots,u^{\ell_0^m}\Big),$$
perform a sequence of minimization
\begin{equation}
    \label{eq:multi_rollout_tau}
        \begin{aligned}
    \Tilde{u}_0^{\ell_0^1}&\in\arg\min_{u^{\ell_0^1}\in U^{{\ell_0^1}}(x)}\Big\{g\big(x,u^{\ell_0^1},\mu^{\ell_0^2}(x),\dots,\mu^{\ell_0^m}(x)\big)\\
    &+\alpha J_\mu\Big(f\big(x,u^{\ell_0^1},\mu^{\ell_0^2}(x),\dots,\mu^{\ell_0^m}(x)\big)\Big)\Big\},\\
    \Tilde{u}_0^{\ell_0^2}&\in\arg\min_{u^{\ell_0^2}\in U^{{\ell_0^2}}(x)}\Big\{g\big(x,\Tilde{u}_0^{\ell_0^1},u^{\ell_0^2},\mu^{\ell_0^3}(x),\dots,\mu^{\ell_0^m}(x)\big)\\
    &+\alpha J_\mu\Big(f\big(x,\Tilde{u}_0^{\ell_0^1},u^{\ell_0^2},\mu^{\ell_0^3}(x),\dots,\mu^{\ell_0^m}(x)\big)\Big\},\\
    \dots& \qquad \dots\qquad \dots\\
    \Tilde{u}_0^{\ell_0^m}&\in\arg\min_{u^{\ell_0^m}\in U^{{\ell_0^m}}(x)}\Big\{g\big(x,\Tilde{u}_0^{\ell_0^1},\dots,\Tilde{u}_0^{\ell_0^{m-1}},u^{\ell_0^m}\big)\\
    &+\alpha J_\mu\Big(f\big(x,\Tilde{u}_0^{\ell_0^1},\dots,\Tilde{u}_0^{\ell_0^{m-1}},u^{\ell_0^m}\big)\Big)\Big\},
        \end{aligned}
\end{equation}
and define a control $\Tilde{u}_0$ as
$$\Tilde{u}_0=\Big(\Tilde{u}_0^{\ell_0^1},\Tilde{u}_0^{\ell_0^2},\dots,\Tilde{u}_0^{\ell_0^m}\Big).$$

If the control $\Tilde{u}_0$ is deemed to be `good' according to the rule, we set $\Tilde{\mu}(x)=\Tilde{u}_0$. Otherwise, we generate according to certain probability distribution a new permutation $\tau_1$, define $\ell_1^i$ similar to \eqref{eq:ell_tau}, and carry out agent-by-agent optimization similar to \eqref{eq:multi_rollout_tau}. This procedure is repeated until after generating, say $j$, new permutations, the corresponding $\Tilde{u}_j$ is considered as `good.' Then we define $\Tilde{\mu}(x)=\Tilde{u}_j$, and the state $x'$ and permutation $\sigma'$ at the next stage is given by
$$x'=f\big(x,\Tilde{\mu}(x)\big),\quad \sigma'=\tau_j.$$

As an example of the rule for selecting controls, let us assume that there exists a function $\hat{J}:X\mapsto\Re$ that sets a realistic bound for the multiagent rollout policy. As stated earlier, the performance bound \eqref{eq:cost_bound} remains valid regardless of the order of agents being optimized. Thus, after obtaining control $\Tilde{u}_j$, we may define a value $\hat{V}_j$ as
$$\hat{V}_j=g(x,\Tilde{u}_j)+\alpha J_\mu\big(f(x,\Tilde{u}_j)\big),$$
and compare $\hat{V}_j$ and $\hat{J}(x)$. If $\hat{V}_j\leq\hat{J}(x)$, then the control $\Tilde{u}_j$ is applied to the system by setting $\Tilde{\mu}(x)=\Tilde{u}_j$. In view of the construction of $\Tilde{\mu}$, we have that 
$$J_{\Tilde{\mu}}(x)\leq \hat{J}(x),\quad \forall x\in X.$$

\section{Modelling for the Warehouse Path Planning Problem}\label{sec:model}
In this section, we present details of the model for the warehouse path finding problem. We will first describe the components of the state $x$ and the control $u$ at each stage, and then introduce the system dynamics $f(\cdot)$ and stage cost $g(\cdot)$ used in our numerical studies presented in the next section. There is a considerable amount of flexibility in the modeling process, and judged by the numerical results, our choices here strike a good balance between implementation expediency and satisfactory performance.

Given that the warehouse environment is modeled as a grid world, we assign to each cell a unique scalar index. We assume that each one of the agents, the goods, as well as the delivery points occupy a single cell. Thus, for $i$th agent, its position at $k$th stage can be described by the index of the cell at which it is located, and is denoted as $p^i_k$. Its current target position is the index of the cell of the good or the delivery point it aims to reach, and is denoted as $t_k^i$. Therefore, information related to the $i$th agent is adequately captured by the pair of indices $(p^i_k,t^i_k)$. All the other information that is varying over stages and is useful for the purpose of defining the dynamics is lumped in a vector denoted as $b_k$. Among others, it includes the positions of goods, which act as obstacles when present. Collecting indices pairs $(p^i_k,t^i_k)$ and $b_k$ together, we define the state at $k$th stage as
$$x_k=\big((p^1_k,t^1_k),(p^2_k,t^2_k),\dots,(p^m_k,t^m_k),b_k\big).$$

 As for control of the $i$th agent, its control constraint set $U^i(x_k)$ at state $x_k$ contains at most $C=5$ elements. They represent the actions of moving to the four neighboring cells located to the left, right, up, and down of its current position $p_k^i$, as well as staying still. Depending on the values of $(p^i_k,t^i_k)$ and $b_k$, some of those controls may be removed due to the boundary of the grid world, as well as other goods acting as obstacles. Therefore, the dependence of set $U^i(x_k)$ on $x_k$ is only through $(p^i_k,t^i_k)$ and $b_k$, but not the positions of other agents. This means that some control $u_k$ that leads to a collision at ($k+1$)th stage (i.e., several agents occupying the same grid at the same time) is considered feasible, namely $u(k)\in U(x_k)$. On the other hand, they are incentivized to avoid such situations by a large cost. Our choice for modeling collision avoidance through cost, rather than control constraints, is mainly motivated by the streamlined mathematical analysis in the preceding section. In practice, there is little difference in the implementation.

The system dynamics $f(\cdot)$ is composed of three parts: mechanisms for updating $p^i_k$, $t^i_k$, and $b_k$ respectively. Given its current position $p^i_k$, $i$th agent applying control $u^i_k$ would lead itself to the position $p^i_{k+1}$ that is either a neighbor of $p^i_k$ or equaling to $p^i_k$ in the case of standstill. Note that we allow the possibility that $p^i_k=p^j_k$ for $i\neq j$, consistent with the feasible control discussed before. As for the target position, we define $t^i_{k+1}=t^i_{k}$ if $p^i_k\neq t^i_k$, meaning that a target is not updated until the target is reached by its corresponding agent. On the other hand, if $p^i_k= t^i_k$, a new target $t^i_{k+1}$ is assigned based upon $x_k$ and $u_k$, the mechanism of which is assumed to be deterministic. In addition, the vector $b_k$ is updated accordingly depending on the remaining goods.  

The stage cost $g(\cdot)$ consists mainly of two parts: a large penalty for collisions, and a negative cost (equivalently, a `reward') for reaching the goal. In particular, the stage cost of applying $u_k$ at $x_k$ is defined as
$$g(x_k,u_k)=g_1(x_k,u_k)+\sum_{i=1}^mg_2(p^i_k,t^i_k),$$
where $g_1(\cdot)$ encodes the collision cost, which is defined as
$$g_1(x_k,u_k)=c_1\cdot n\big(f(x_k,u_k)\big),$$
with $c_1$ being a large positive constant, and $n$ mapping a state to the number of collisions that occurred in the state. The function $g_2(p^i_k,t^i_k)$ takes a negative value $c_2$ if $p^i_k= t^i_k$ and $0$ otherwise. 

At this point, together with a suitably defined discount factor $\alpha$, we obtain a multiagent deterministic optimal control problem with details filled in. The problem is then addressed by the multiagent rollout with reshuffling algorithm introduced in Section~\ref{sec:multi_rollout}. We apply two base policies $\mu_1$ and $\mu_2$, with $\mu_1$ being the shortest paths for each agent computed by A*, while the $\mu_2$ being the same as $\mu_1$ except that it freezes robots for different numbers of steps depending on their indices, i.e., staying still for a few consecutive stages. The rule used to define a `good' control is that a control $u_k\in U(x_k)$ at $x_k$ is considered `good' if $g_1(x_k,u_k)=0$, or equivalently, no immediate collision is caused by applying $u_k$ at $x_k$. The testing results with our algorithm are presented next.     

\section{Numerical Studies}\label{sec:numerical}
This section presents the results of our numerical studies. The environment used here is composed of $47\times 115$ grids, whose layout resembles that of \citep[Fig.~3]{wurman2008coordinating}. There are $m$ agents which are tasked to collaboratively deliver $1183$ goods to their designated delivery points. At any given time, if the number of remaining goods is larger than $m$, each robot is assigned randomly a good, whose delivery point is also chosen randomly.\footnote{Throughout our implementation, all random events are generated according to a uniform distribution.} If the scheduling mechanism is available, this random procedure can be duly replaced. A new good among the remaining ones is assigned to a robot once its current task is completed. Note that our approach applies to the life-long delivery mission, by which we mean that there are infinite amounts of goods arriving sequentially for delivery. For convenience, we restrict our attention to a finite goods case. A test where all the goods have been delivered successfully or where a collision has occurred is named an `episode.' The problem data of the optimal control problem described in Section~\ref{sec:model} are $\alpha=0.999$, $c_1=10^{20}$, and $c_2=-10^4$.

We use cooperative A* introduced in \citep{silver_cooperative_2005} as a benchmark for comparison (`CA*' for short in the figures). It computes the shortest paths of agents according to a fixed order via A* in a space-time grid, while treating the paths of preceding agents as obstacles. As the scheme is a one-short algorithm (see the discussion in Section~\ref{sec:intro}), it is executed every time a new assignment is given (or equivalently, a delivery is completed, provided that there are remaining goods to be assigned). Since simulation in our scheme deals with a static environment, we can compute all the shortest paths to all goods and delivery points offline and store them for online simulation. This procedure is called pre-computation. Our scheme with and without pre-computation is labeled as `MA-rollout' and `MA-rollout w/o pre-comp' in the figures respectively. For those approaches, we test cases with agent number $m$ ranging from $100$ to $200$ in the same environment, and the listed results are averaged over successful tests among 200 episodes for each agent count. An Apple M1 processor was used for all tests, which were performed on a single thread. The detailed implementation of our scheme as well as demonstrations of a few test cases can be found in \url{https://github.com/will-em/multi-agent-rollout}.

\subsection{Success rate, computation time, and memory usage}
\begin{figure}
    \centering
    \begin{tikzpicture}[trim axis left, trim axis right]
    \begin{axis}[
        xlabel={Number of agents $m$},
        ylabel={Success rate [\%]},
        xmin=100, xmax=200,
        ymin=50, ymax=100,
        width=7.5cm,
        height=4cm,
        xtick={100,150,200},
        ytick={},
        legend pos=south west,
        ymajorgrids=true,
        grid style=dashed,
        legend style={nodes={scale=0.6, transform shape}},
    ]
    
    \addplot[
        plotline1,
        ]
        coordinates {
        (100, 89)(110,90)(120,89)(130,84)(140,80)(150,79)(160, 72)(170, 76)(180, 67)(190, 67)(200, 58)
        };
    \addlegendentry{CA*}
    
    \addplot[
        plotline2,
        ]
        coordinates {
        (100, 100)(110,100)(120,100)(130,100)(140,100)(150,100)(160, 100)(170, 100)(180, 100)(190, 100)(200, 100)
        };
    \addlegendentry{MA-rollout}

    \end{axis}
    \end{tikzpicture}
        \caption{The success rate (in percent) at which the methods are able to complete the task.}
        \vspace{2mm}
        \label{fig:success_rate}
\end{figure}
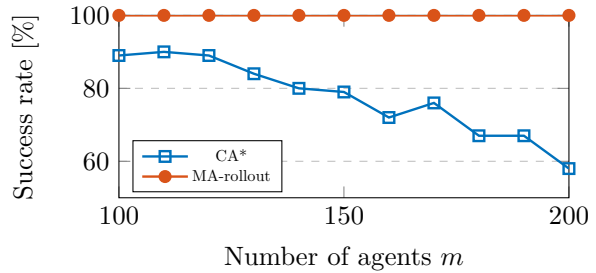

\begin{figure}
    \centering

    \begin{tikzpicture}[trim axis left, trim axis right]
    \begin{axis}[
        xlabel={Number of agents $m$},
        ylabel={$J_{\Tilde{\mu}}(x_0)$},
        xmin=100, xmax=200,
        ymin=-15000000, ymax=-10000000,
        width=7.5cm,
        height=4cm,
        xtick={100,150,200},
        ytick={},
        legend pos=north east,
        ymajorgrids=true,
        grid style=dashed,
        legend style={nodes={scale=0.6, transform shape}},
    ]
    
    \addplot[
        plotline1,
        ]
        coordinates {
        (100, -11067000)(110,-11633200)(120,-12114500)(130,-12573800)(140,-12974900)(150,-13340400)(160, -13638800)(170, -13864100)(180, -14086000)(190, -14284500)(200, -14465500)
        };
    \addlegendentry{CA*}
    
    \addplot[
        plotline2,
        ]
        coordinates {
        (100, -10375300)(110,-10891900)(120,-11368100)(130,-11711800)(140,-11964700)(150,-12174700)(160, -12347200)(170, -12464300)(180, -12631600)(190, -12755200)(200, -12825800)
        };
    \addlegendentry{MA-rollout}

    \end{axis}
    \end{tikzpicture}
        \caption{The cumulative cost $J_{\Tilde{\mu}}$, averaged over 200 episodes. This is far smaller than $\Tilde{J}$, which has an average value of around $10^{20}$, verifying inequality \eqref{eq:cost_bound}.}
        \vspace{2mm}
        \label{fig:cumulative_reward}
\end{figure}
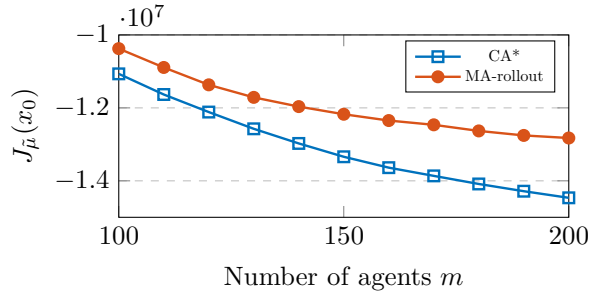

The rates of successful episodes with our scheme and cooperative A* are shown in Fig.~\ref{fig:success_rate}. Our approach is able to complete all delivery missions without any collision occurring in all tests with on average less than $2$ reshuffling procedures, while the success rate drops to about $60\%$ when $m=200$ in the case of cooperative A*. Note that for $m=100$, our result of cooperative A* is consistent with the result reported in the original work \citep[Fig.~3]{silver_cooperative_2005}. Among those where cooperative A* completes the task, its average cost is lower than that of our approach, as shown in Fig.~\ref{fig:cumulative_reward}. Still, both performance bounds \eqref{eq:cost_bound} and \eqref{eq:rollout_improve} for our scheme are verified in those tests.

\vspace{-0.1cm}

\vspace{-0.1cm}

\begin{figure}
    \centering

    \begin{tikzpicture}[trim axis left, trim axis right]
    \begin{axis}[
        xlabel={Number of agents $m$},
        ylabel={Runtime [ms]},
        xmin=100, xmax=200,
        ymin=5, ymax=60,
        width=7.5cm,
        height=4cm,
        xtick={100,150,200},
        ytick={10, 20, 30, 40, 50},
        legend pos=north west,
        ymajorgrids=true,
        grid style=dashed,
        legend style={nodes={scale=0.65, transform shape}},
    ]
    
    \addplot[
        plotline1,
        ]
        coordinates {
        (100, 19.0847)(110,20.6146)(120,22.2105)(130,24.2083)(140,25.9375)(150,27.9247)(160, 29.4)(170, 31.1875)(180, 32.5426)(190, 34.8866)(200, 38.5506)
        };
    \addlegendentry{CA*}
        
    \addplot[
        plotline2,
        ]
        coordinates {
        (100, 8.35895)(110,10.3783)(120,12.8847)(130,15.2359)(140,18.2616)(150,22.0642)(160, 25.5706)(170, 30.7799)(180, 35.61)(190, 42.7839)(200, 47.7844)
        };
    \addlegendentry{MA-rollout}
    
    \addplot[
        plotline3,
        ]
        coordinates {
        (100, 9.95)(110,11.93)(120,14.31)(130,16.7216)(140,20.5228)(150,24.1993)(160, 28.6255)(170, 33.6049)(180, 38.0182)(190, 45.5216)(200, 54.16)
        };
    \addlegendentry{MA-rollout w/o pre-comp.}

    \end{axis}
    \end{tikzpicture}
        \caption{The average computation time required to obtain feasible controls for all agents.}
        \label{fig:runtime}
\end{figure}
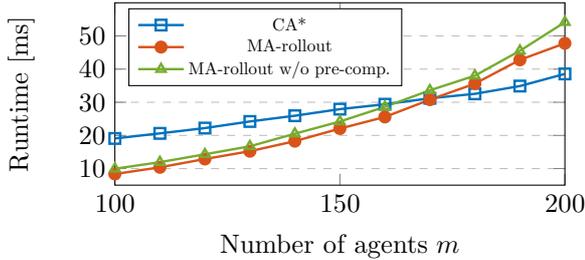

The computation times for different approaches are averaged over their respective successful episodes and shown in Fig.~\ref{fig:runtime}. Although cooperative A* has a slight edge over our method when $m>160$, its successful episodes are less than $80\%$ of the total tests. In addition, similar to the generic rollout method, our scheme is well-suited for parallel computation. Therefore, the computation time for our scheme would be around $1/C$ of what is shown in Fig.~\ref{fig:runtime} had parallel computation been fully brought to bear, see \citep{bertsekas2021multiagent} for further details. The pre-computation of shortest paths reduces about $20\%$ computation time in our tests.

Besides, the scale of the graphs where the shortest-path searches are performed differs between the methods. The graph needed by cooperative A* is expected to be $N$ times larger than that of our scheme, where $N$ denotes the largest number of stages needed for all agents to reach their current respective targets. This may therefore pose limitations on the maximum path length of cooperative A* from a memory standpoint. For the pre-computation used in our scheme, the total size of the data is the dimension of the environment times the sum of the number of the spots where the goods are placed and that of the delivery points, which in our example is around $7$ Mb.

\subsection{Adaptivity}
Since the real-world operation of a warehouse is prone to unforeseeable events, it is of great importance for a warehouse path-finding algorithm to be able to adapt to these circumstances. The proposed algorithm can adapt to a changing environment by means of online replanning, and we demonstrate this ability by addressing randomly occurring robot malfunction, which is modeled by using a failure mode for the robots. In particular, up to $20\%$ of robots are fixed at their position indefinitely starting from a random stage and are acting as obstacles ever since. Their assigned tasks are later assigned to the functioning robots. The indices of malfunctioning robots are assumed known. To cope with this situation, our proposed scheme simply regards the malfunctioned robots as the ones with modified control constraint sets $\hat{U}^i(x_k)$, containing nothing but one control that corresponds to `staying still.' Then our scheme can be applied without any further modification. We carried out 50 tests with $100$ agents. Our scheme adapted to this kind of scenario, and the remaining agents continued their operation by circumnavigating these malfunctioned robots.

\section{Conclusion}
We proposed the multiagent rollout with reshuffling for solving the warehouse robot path planning problem. Based upon the framework introduced in \citep{bertsekas2021rollout}, our scheme can produce paths for a large number of robots in real time while achieving collision avoidance. Unlike existing schemes that are searching in environments associated with both space and time, our method relies on the use of simulation in a smaller static environment. It also inherits the generic features of rollout and can adapt to a changing environment by means of online replanning. Our numerical studies demonstrated that the scheme scaled well with a large number of robots and could adapt to situations where robot malfunctions occurred.  

\begin{ack}
The authors are grateful to Prof. Dimitri P. Bertsekas for valuable suggestions and comments in various stages of the project. They also appreciate the contributions of Laura Briffa in implementing early versions of the scheme. 
\end{ack}

\bibliography{ifacconf}

\begin{thebibliography}{20}
\providecommand{\natexlab}[1]{#1}
\providecommand{\url}[1]{\texttt{#1}}
\providecommand{\urlprefix}{URL }
\expandafter\ifx\csname urlstyle\endcsname\relax
  \providecommand{\doi}[1]{doi:\discretionary{}{}{}#1}\else
  \providecommand{\doi}{doi:\discretionary{}{}{}\begingroup
  \urlstyle{rm}\Url}\fi

\bibitem[{Bellman(1958)}]{bellman_routing_1958}
Bellman, R. (1958).
\newblock On a routing problem.
\newblock \emph{Quarterly of Applied Mathematics}, 16(1), 87--90.

\bibitem[{Bertsekas(2019)}]{bertsekas2019reinforcement}
Bertsekas, D. (2019).
\newblock \emph{Reinforcement learning and optimal control}.
\newblock Athena Scientific.

\bibitem[{Bertsekas(2021{\natexlab{a}})}]{bertsekas2021multiagent}
Bertsekas, D. (2021{\natexlab{a}}).
\newblock Multiagent reinforcement learning: Rollout and policy iteration.
\newblock \emph{IEEE/CAA Journal of Automatica Sinica}, 8(2), 249--272.

\bibitem[{Bertsekas(2021{\natexlab{b}})}]{bertsekas2021rollout}
Bertsekas, D. (2021{\natexlab{b}}).
\newblock \emph{Rollout, policy iteration, and distributed reinforcement
  learning}.
\newblock Athena Scientific.

\bibitem[{Bertsekas and Tsitsiklis(1996)}]{bertsekas1996neuro}
Bertsekas, D. and Tsitsiklis, J.N. (1996).
\newblock \emph{Neuro-dynamic programming}.
\newblock Athena Scientific.

\bibitem[{Bertsekas(1991)}]{bertsekas1991auction}
Bertsekas, D.P. (1991).
\newblock An auction algorithm for shortest paths.
\newblock \emph{SIAM Journal on Optimization}, 1(4), 425--447.

\bibitem[{Bhattacharya et~al.(2021)Bhattacharya, Kailas, Badyal, Gil, and
  Bertsekas}]{bhattacharya2021multiagent}
Bhattacharya, S., Kailas, S., Badyal, S., Gil, S., and Bertsekas, D. (2021).
\newblock Multiagent rollout and policy iteration for pomdp with application to
  multi-robot repair problems.
\newblock In \emph{Conference on Robot Learning}, 1814--1828. PMLR.

\bibitem[{{\v{C}}{\'a}p et~al.(2015){\v{C}}{\'a}p, Nov{\'a}k, Kleiner, and
  Seleck{\`y}}]{vcap2015prioritized}
{\v{C}}{\'a}p, M., Nov{\'a}k, P., Kleiner, A., and Seleck{\`y}, M. (2015).
\newblock Prioritized planning algorithms for trajectory coordination of
  multiple mobile robots.
\newblock \emph{IEEE transactions on automation science and engineering},
  12(3), 835--849.

\bibitem[{Dijkstra(1959)}]{dijkstra_note_1959}
Dijkstra, E.W. (1959).
\newblock A note on two problems in connexion with graphs.
\newblock \emph{Numerische Mathematik}, 1(1), 269--271.

\bibitem[{Hart et~al.(1968)Hart, Nilsson, and Raphael}]{hart_formal_1968}
Hart, P., Nilsson, N., and Raphael, B. (1968).
\newblock A {Formal} {Basis} for the {Heuristic} {Determination} of {Minimum}
  {Cost} {Paths}.
\newblock \emph{IEEE Transactions on Systems Science and Cybernetics}, 4(2),
  100--107.

\bibitem[{Latombe(1991)}]{latombe2012robot}
Latombe, J.C. (1991).
\newblock \emph{Robot motion planning}, volume 124.
\newblock Springer Science \& Business Media.

\bibitem[{Li et~al.(2021)Li, Tinka, Kiesel, Durham, Kumar, and
  Koenig}]{li_lifelong_2021}
Li, J., Tinka, A., Kiesel, S., Durham, J.W., Kumar, T.K.S., and Koenig, S.
  (2021).
\newblock Lifelong {Multi}-{Agent} {Path} {Finding} in {Large}-{Scale}
  {Warehouses}.

\bibitem[{Ma et~al.(2017)Ma, Li, Kumar, and Koenig}]{ma_lifelong_2017}
Ma, H., Li, J., Kumar, T.S., and Koenig, S. (2017).
\newblock Lifelong {Multi}-{Agent} {Path} {Finding} for {Online} {Pickup} and
  {Delivery} {Tasks}.
\newblock In \emph{Proceedings of the 16th {Conference} on {Autonomous}
  {Agents} and {MultiAgent} {Systems}}, {AAMAS} '17, 837--845. International
  Foundation for Autonomous Agents and Multiagent Systems, Richland, SC.

\bibitem[{Sharon et~al.(2015)Sharon, Stern, Felner, and
  Sturtevant}]{sharon_conflict-based_2015}
Sharon, G., Stern, R., Felner, A., and Sturtevant, N.R. (2015).
\newblock Conflict-based search for optimal multi-agent pathfinding.
\newblock \emph{Artificial Intelligence}, 219, 40--66.
\newblock \doi{10.1016/j.artint.2014.11.006}.

\bibitem[{Silver(2005)}]{silver_cooperative_2005}
Silver, D. (2005).
\newblock Cooperative {Pathfinding}.
\newblock In \emph{Proceedings of the AAAI conference on artificial
  intelligence and interactive digital entertainment}, 117--122.

\bibitem[{Silver et~al.(2018)Silver, Hubert, Schrittwieser, Antonoglou, Lai,
  Guez, Lanctot, Sifre, Kumaran, Graepel et~al.}]{silver2018general}
Silver, D., Hubert, T., Schrittwieser, J., Antonoglou, I., Lai, M., Guez, A.,
  Lanctot, M., Sifre, L., Kumaran, D., Graepel, T., et~al. (2018).
\newblock A general reinforcement learning algorithm that masters chess, shogi,
  and go through self-play.
\newblock \emph{Science}, 362(6419), 1140--1144.

\bibitem[{Sutton and Barto(2018)}]{sutton2018reinforcement}
Sutton, R.S. and Barto, A.G. (2018).
\newblock \emph{Reinforcement learning: An introduction}.
\newblock MIT press.

\bibitem[{Tesauro and Galperin(1996)}]{tesauro1996line}
Tesauro, G. and Galperin, G. (1996).
\newblock On-line policy improvement using monte-carlo search.
\newblock In \emph{NeurIPS}.

\bibitem[{Wagner and Choset(2011)}]{wagner_m_2011}
Wagner, G. and Choset, H. (2011).
\newblock M*: {A} complete multirobot path planning algorithm with performance
  bounds.
\newblock In \emph{2011 {IEEE}/{RSJ} {IROS}}, 3260--3267.
\newblock \doi{10.1109/IROS.2011.6095022}.

\bibitem[{Wurman et~al.(2008)Wurman, D'Andrea, and
  Mountz}]{wurman2008coordinating}
Wurman, P.R., D'Andrea, R., and Mountz, M. (2008).
\newblock Coordinating hundreds of cooperative, autonomous vehicles in
  warehouses.
\newblock \emph{AI magazine}, 29(1), 9--9.

\end{thebibliography}

\appendix
\section{Performance bound of multiagent rollout}\label{app:bound}
We state and prove a proposition that formally establishes the performance bound given in \eqref{eq:cost_bound}. 

\begin{prop}[Performance bound of multiagent rollout]\label{prop:perform_bound}
Consider the multiagent rollout policy $\Tilde{\mu}$ defined by \eqref{eq:multi_rollout}. The function $\Tilde{J}:X\mapsto \Re$ given in \eqref{eq:j_tilde} is an upper bound of the cost function of policy $\Tilde{\mu}$, i.e.,
$$J_{\Tilde{\mu}}(x)\leq \Tilde{J}(x),\quad \forall x\in X.$$
\end{prop}

To prove the above proposition, we first recall some classical results, which hold well beyond the scope of the problems considered here. The form given below is adopted from \citep[Prop.~4.3.3]{bertsekas2019reinforcement}.\footnote{Note that the statement given in \citep[Prop.~4.3.3]{bertsekas2019reinforcement} uses the transition probability format, while here we use system dynamics form. This is because the problem considered here is deterministic, thus making the transition probability format cumbersome. Still, those two formats are equivalent, and one may refer to \citep{bertsekas2019reinforcement} p.~177 for a justification of the equivalence.}

\begin{lem}\label{lem:classic}
For arbitrary policy $\mu$, and functions $J$ and $J'$ that map $X$ to $\Re$:
\begin{itemize}
    \item[(a)] If $J(x)\leq J'(x)$ for all $x\in X$, then 
    \begin{align*}
        &g\big(x,\mu(x)\big)+\alpha J\Big(f\big(x,\mu(x)\big)\Big)\\
        \leq &g\big(x,\mu(x)\big)+\alpha J'\Big(f\big(x,\mu(x)\big)\Big),\quad \forall x\in X.
    \end{align*}
    \item[(b)] We have that 
    \begin{equation}
        \label{eq:bellman_mu}
        J_{\mu}(x)=g\big(x,\mu(x)\big)+\alpha J_\mu\Big(f\big(x,\mu(x)\big)\Big),\quad \forall x\in X.
    \end{equation}
    \item[(c)] Let $J_0(x)=J(x)$ for all $x$, and consider the sequence $\{J_k\}$ generated according to the iteration
    \begin{equation}
        \label{eq:vi}
        J_{k+1}(x)=g\big(x,\mu(x)\big)+\alpha J_k\Big(f\big(x,\mu(x)\big)\Big),\quad \forall x\in X.
    \end{equation}
    Then the generated sequence converges to the cost function of policy $\mu$, i.e.,
    $$\lim_{k\to\infty}J_k(x)=J_\mu(x).$$
\end{itemize}
\end{lem}
For the convenience of reference, in what follows, we will refer to Lemma~\ref{lem:classic}(a) as the `monotonicity' property, while referring to the algorithm defined in \eqref{eq:vi} as VI of $\mu$.

\begin{pf}[Proof of Prop.~\ref{prop:perform_bound}]
First, in view of the construction of $\Tilde{\mu}$ and the definition \eqref{eq:j_tilde} of $\Tilde{J}$, we have that
\begin{align*}
    \Tilde{J}(x)=&g\big(x,\Tilde{\mu}(x)\big)+\alpha J_\mu\Big(f\big(x,\Tilde{\mu}(x)\big)\Big)\\
    \leq&g\big(x,\mu(x)\big)+\alpha J_\mu\Big(f\big(x,\mu(x)\big)\Big).
\end{align*}
Combining above inequality with the equation \eqref{eq:bellman_mu}, we have that
\begin{equation}
    \label{eq:j_tilde_j_mu}
    \Tilde{J}(x)\leq J_\mu(x),\quad \forall x\in X.
\end{equation}

Then we consider the VI of $\Tilde{\mu}$ starting from $J_\mu$, i.e., $J_0(x)=J_\mu(x)$ for all $x$, and 
$$J_{k+1}(x)=g\big(x,\Tilde{\mu}(x)\big)+\alpha J_k\Big(f\big(x,\Tilde{\mu}(x)\big)\Big),\quad \forall x\in X.$$
Clearly, we have $\Tilde{J}(x)=J_1(x)$ for all $x$. In view of \eqref{eq:j_tilde_j_mu}, and applying the monotonicity property with $\Tilde{J}$ and $J_\mu$ in place of $J$ and $J'$, and $\Tilde{\mu}$ in place of $\mu$, we have that
\begin{align*}
    J_2(x)=&g\big(x,\Tilde{\mu}(x)\big)+\alpha J_1\Big(f\big(x,\Tilde{\mu}(x)\big)\Big)\\
    =&g\big(x,\Tilde{\mu}(x)\big)+\alpha \Tilde{J}\Big(f\big(x,\Tilde{\mu}(x)\big)\Big)\\
        \leq &g\big(x,\Tilde{\mu}(x)\big)+\alpha J_\mu\Big(f\big(x,\Tilde{\mu}(x)\big)\Big)\\
        =&\Tilde{J}(x)=J_1(x),\quad \forall x\in X.
\end{align*}
Repeating the above steps yield 
$$J_{k+1}(x)\leq J_k(x),\quad\forall x\in X,\;k=0,1,\dots.$$
In addition, due to the convergence property stated in Lemma~\ref{lem:classic}(c), we conclude that 
$$J_{\Tilde{\mu}}(x)=\lim_{k\to\infty}J_k(x)\leq J_1(x)=\Tilde{J}(x),\quad \forall x\in X.$$
\end{pf}                                        
\end{document}